\documentclass{article}
\usepackage[T1]{fontenc} 
\usepackage[utf8]{inputenc} 
\usepackage{ismir,amsmath,cite,url}
\usepackage{graphicx}
\usepackage{color}
\usepackage{multirow}
\usepackage{multicol}
\usepackage{booktabs}
\usepackage{colortbl} 
\usepackage{bbding}
\usepackage{bbm}
\definecolor{mygray}{gray}{0.6}
\definecolor{mygray-bg}{gray}{0.95}
\usepackage{lineno}
\usepackage{url}

\title{LyricWhiz: Robust Multilingual Zero-shot Lyrics Transcription by Whispering to ChatGPT}

\multauthor
{\small Le Zhuo$^1$ \hspace{0.2cm} Ruibin Yuan$^{2,3}$ \hspace{0.2cm} Jiahao Pan$^4$ \hspace{0.2cm} Yinghao Ma$^5$ \hspace{0.2cm} Yizhi Li$^6$ \hspace{0.2cm} Ge Zhang$^{2,7}$ \hspace{0.2cm} Si Liu$^1$} 
{\bfseries{\small Roger Dannenberg$^3$ \hspace{0.2cm} Jie Fu$^2$ \hspace{0.2cm} Chenghua Lin$^6$ \hspace{0.2cm} Emmanouil Benetos$^5$ \hspace{0.2cm} \hspace{0.2cm} Wei Xue$^4$ \hspace{0.2cm} Yike Guo$^4$}\\
\small
    $^1$ Beihang University \hspace{0.2cm}
    $^2$ Beijing Academy of Artificial Intelligence \hspace{0.2cm}
    $^3$ Carnegie Mellon University\\
\small
    $^4$ Hong Kong University of Science and Technology \hspace{0.2cm}
    $^5$ Queen Mary University of London\\
\small
    $^6$ University of Sheffield \hspace{0.2cm}
    $^7$ University of Waterloo\\
{\tt\small zhuole1025@gmail.com, ruibiny@andrew.cmu.edu, fujie@baai.ac.cn}
}

\def\authorname{L Zhuo, R Yuan, and J Pan}

\begin{document}

\maketitle

\begin{abstract}
We introduce LyricWhiz, a robust, multilingual, and zero-shot automatic lyrics transcription method achieving state-of-the-art performance on various lyrics transcription datasets, even in challenging genres such as rock and metal.
Our novel, training-free approach utilizes Whisper, a weakly supervised robust speech recognition model, and GPT-4, today's most performant chat-based large language model. In the proposed method, Whisper functions as the ``ear'' by transcribing the audio, while GPT-4 serves as the ``brain,'' acting as an annotator with a strong performance for contextualized output selection and correction.
Our experiments show that LyricWhiz significantly reduces Word Error Rate compared to existing methods in English and can effectively transcribe lyrics across multiple languages.
Furthermore, we use LyricWhiz to create the first publicly available, large-scale, multilingual lyrics transcription dataset with a CC-BY-NC-SA copyright license, based on MTG-Jamendo, and offer a human-annotated subset for noise level estimation and evaluation. We anticipate that our proposed method and dataset will advance the development of multilingual lyrics transcription, a challenging and emerging task. The code and dataset are available at \url{https://github.com/zhuole1025/LyricWhiz}.
\end{abstract}

\section{Introduction}\label{sec:introduction}

\begin{figure}[h]
    \centering
    \includegraphics[width=.95\linewidth]{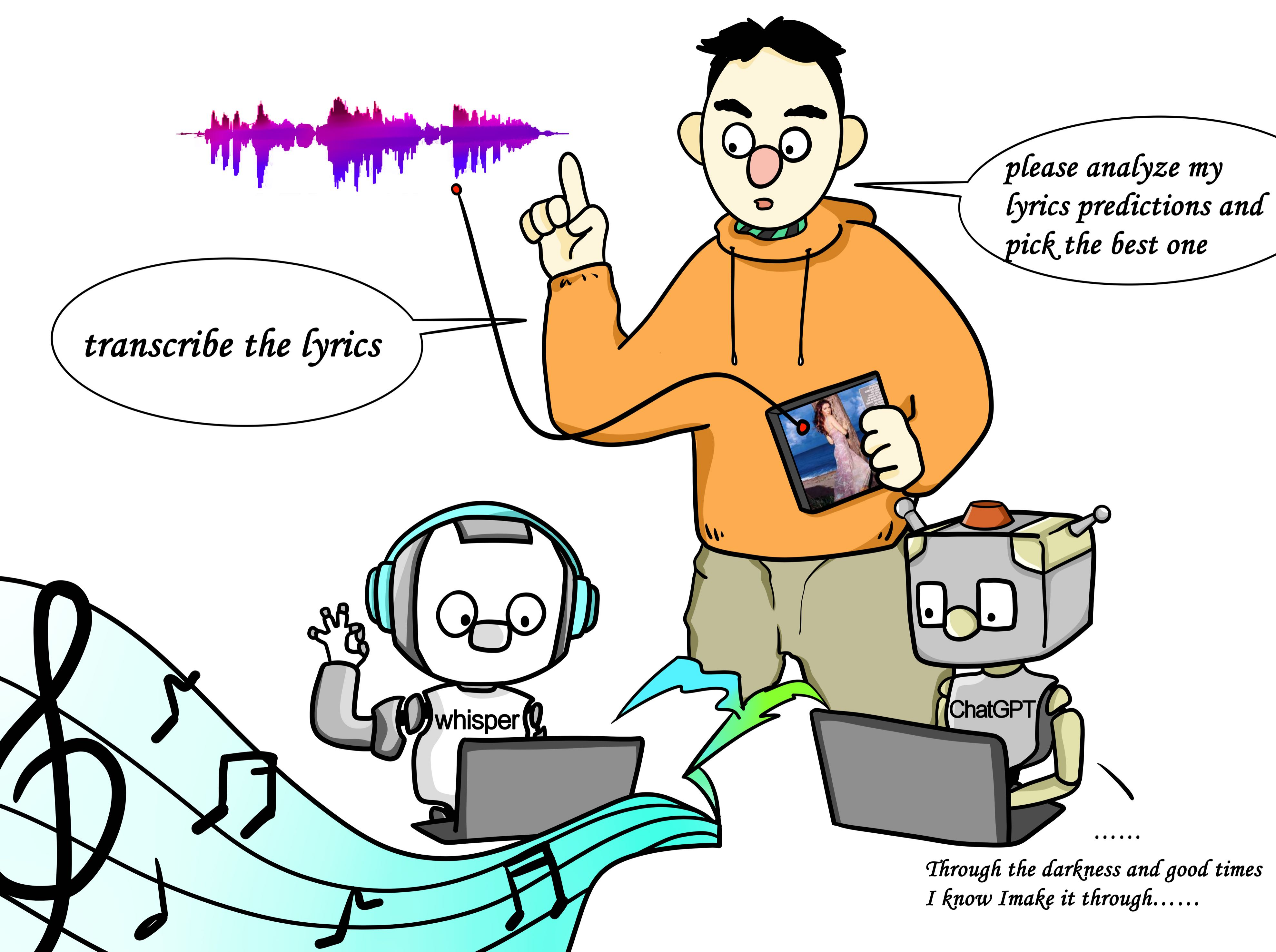}
    \caption{Concept illustration of the working LyricWhiz, where user prompts the two advanced models, Whisper and ChatGPT, to perform automatic lyrics transcription.}\label{fig:fun}
    \vspace{-4mm}
\end{figure}

Automatic lyrics transcription (ALT) is a crucial task in music information retrieval (MIR) that involves converting an audio recording into a textual representation of the lyrics sung in the recording. 
The importance of this task stems from the fact that lyrics are a fundamental aspect of many music genres and are often the main way in which listeners engage with and interpret a song's meaning. 
Additionally, ALT has numerous applications in the music industry, such as enabling better cataloging~\cite{tsaptsinos2017lyrics}, music searching~\cite{fujihara2008hyperlinking,hosoya2005lyrics}, music recommendation~\cite{knees2013survey}, as well as facilitating the creation of karaoke tracks and lyric videos. Moreover, ALT can assist in various music-related research tasks, including sentiment analysis~\cite{ccano2017moodylyrics}, music genre classification~\cite{tsaptsinos2017lyrics}, lyrics generation, which is further used for music generation~\cite{dhariwal2020jukebox}, security review, and music copyright protection. Thus, accurate and efficient ALT is essential for advanced MIR and the development of new music-related applications.

However, to date, no sufficiently robust and accurate ALT system has been developed. Even major commercial music streaming platforms still rely heavily on manually-annotated lyrics, incurring high costs. One key reason is the challenging nature of lyrics transcription. The diversity of singing styles and skills leads to varied timbres of the same pronunciation. Moreover, the phonemes in singing may be pronounced in vastly different ways, such as longer duration, tone changes, or even vowel substitutions, to accommodate the melody. Lastly, the inclusion of various music accompaniments across different genres makes it challenging to distinguish the vocal signals from other sounds. To surmount these challenges, a more robust ALT system is necessary, capable of outperforming existing models in diverse scenarios, including the transcription of multilingual lyrics.

Another significant factor hindering the progress of ALT systems is the absence of large-scale singing datasets. Currently, only two relatively sizable datasets~\cite{meseguer2019dali, dabike2019automatic} exist for ALT systems. However, all existing datasets are in English, with no multilingual datasets available. Besides, these datasets often have stringent copyright licensing restrictions, which significantly hampers their utilization by researchers. Consequently, developing a more comprehensive and representative dataset, encompassing multiple languages and without copyright issues, is essential for supporting the creation of a robust and accurate system.

In this paper, we present LyricWhiz, a novel method for automatic lyrics transcription. LyricWhiz surpasses existing methods on various ALT datasets, resulting in a significant reduction in WER for English lyrics and providing accurate transcription results across multiple languages. Our system is robust, multilingual, and training-free. To achieve these results, we combined two powerful models from their respective domains as shown in~\figref{fig:fun}: Whisper, a weakly supervised speech transcription model, and GPT-4, a large language model (LLM) from the ChatGPT family. Whisper acts as the ``ear'' while GPT-4 serves as the ``brain'' by providing contextualized output selection and correction with strong performance~\cite{törnberg2023chatgpt4}. We further use LyricWhiz to build a multilingual lyrics dataset, named MulJam, which is the first large-scale, multilingual lyrics transcription dataset without copyright-related issues.

The contributions of our work are as follows:
\begin{itemize}
\item We propose a novel, robust, training-free ALT  method, LyricWhiz, which significantly reduces WER on various ALT benchmark datasets, including Jamendo, Hansen, and MUSDB18, and is close to the in-domain state-of-the-art system on DSing.
\item We introduce the first ALT system that can perform zero-shot, multilingual, long-form ALT by integrating a large speech transcription model and an LLM for contextualized post-processing.
\item We create the first publicly-available, large-scale, multilingual lyrics transcription dataset with a clear copyright statement which eliminates further reviewing of the users and facilitates public usage. We provide a human-annotated subset to estimate noise levels and evaluate multilingual ALT performance.
\end{itemize}

\begin{figure*}[h]
    \centering
    \includegraphics[width=0.8\linewidth]{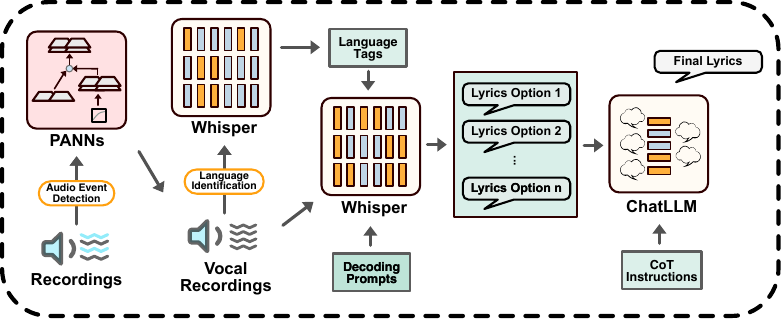}
    \caption{Framework of the proposed LyricWhiz. In the first stage, we employ PANNS~\cite{kong2020panns}, to detect audio events and filter out non-vocal recordings. In the second stage, we utilize the language identification module in Whisper to predict input audio language. We then construct language-specific prompts for Whisper and transcribe input audio multiple times. In the final stage, we request ChatGPT with CoT instructions to ensemble multiple predictions and generate the final lyrics.}\label{fig:framework}
    \vspace{-3mm}
\end{figure*}

\vspace{-2mm}
\section{Related Work}\label{sec:related}
\subsection{Automatic Lyrics Transcription}
Automatic lyrics transcription (ALT) is an essential task in music information retrieval and analysis, aiming to recognize lyrics from singing voices. It remains challenging due to facts such as the sparsity of training data and the unique acoustic characteristics of the singing voice that differ from normal speech. Traditional methods treat ALT in the automatic speech recognition (ASR) framework, which generally utilizes a hybrid of language model and acoustic model, e.g., HMM-GMM. 
Music-related characteristics have been used to further address these challenges~\cite{gupta2018automatic,gupta2020automatic,demirel2021mstre}. 

Despite integrating domain-specific music priors into system designs, the data scarcity issue persists. 
Recently, some researchers have constructed datasets for end-to-end learning, which greatly advances ALT, but most datasets are either noisy (DALI~\cite{meseguer2019dali,meseguer2020creating}, Hansen \cite{hansen2012recognition}, 
DAMP-MVP\footnote{https://zenodo.org/record/2747436\#.ZDqBQOzML0o}); not large (Vocadito \cite{bittner2021vocadito}); or not diverse in terms of genre and language (MUSDB18~\cite{schulze2021phoneme}, DSing~\cite{dabike2019automatic}).

Recent rapid progress in ASR has greatly benefited ALT. Some work focuses on applying the ASR model architectures~\cite{gao2022genre,basak2021end,zhang2021pdaugment}, such as the Transformers, to ALT, and other work leverages the vast amount of public annotated ASR datasets~\cite{kruspe2015training,basak2021end,zhang2021pdaugment} to bridge between the speech and music data. For the first time, a recent study~\cite{ou2022transfer} transferred a large-scale self-supervised pre-trained ASR model, mus2vec 2.0, to the singing domain, and exhibited superior performance on multiple benchmark datasets. Nevertheless, this approach consists of pre-training, fine-tuning, and transfer learning phases, thereby remaining relatively complicated and still requiring singing datasets.


\subsection{Weakly Supervised Automatic Speech Recognition}
The paradigm of large-scale unsupervised pretraining and non-large annotated dataset finetuning has dominated end-to-end ASR research~\cite{tang2022speechnet}. Well-known pretrained ASR models include contrastive learning based Wav2vec~\cite{schneider2019wav2vec}, Wav2vec 2.0~\cite{baevski2020wav2vec},
HuBert~\cite{hsu2021hubert}, WavLM~\cite{chen2022wavlm}, 
Whisper~\cite{radford2022robust}, and Vall-E~\cite{wang2023neural}, which have performed impressively in various downstream tasks, including ASR and speech synthesis. Among them, Whisper has been most recognized for its ASR robustness across different datasets and its multilingual and multitasking capabilities, making Whisper potentially applicable to music tasks. Besides, specifically for ALT, pre-trained musical audio models including JukeBox~\cite{dhariwal2020jukebox}, MusicLM~\cite{agostinelli2023musiclm}, MULE~\cite{mccallum2022supervised}, SingSong~\cite{donahue2023singsong}, music2vec~\cite{li2022map}, and MERT~\cite{li2023mert}, may also contribute to achieving strong performance.

\subsection{Chat-based Large Language Models}
ChatGPT~\footnote{https://openai.com/blog/chatgpt}, a chat-based large language model (LLM), has found broad application in optimizing workflows across a variety of domains, including multimodal intelligence~\cite{shen2023hugginggpt,Schick2023ToolformerLM}.
Recent breaking AutoGPT~\footnote{https://github.com/Torantulino/Auto-GPT} is even recognized as an embryonic form of artificial general intelligence. Inspired by these developments, LyricWhiz collaborates with both Whisper~\cite{radford2022robust} and ChatGPT to optimize the workflow of ALT. 
Prompt engineering is known to be important to navigate LLMs to perform better~\cite{liu2023pre}. LyricWhiz mainly adopts three primary strategies:

a) As shown in~\cite{white2023chatgpt, white2023prompt}, a well-formalized task description prompt can effectively improve ChatGPT's performance on downstream tasks with strict format requirements.
We follow this empirical observation to strictly formalize the expected format of ALT post-processing outputs.
We also refer to the prompt pattern catalog in~\cite{white2023prompt} for an intuitive understanding of prompt engineering.

b) Inspired by \cite{shin2020autoprompt, shum2023automatic}, LyricWhiz utilizes prompt augmentation to ask ChatGPT to analyze the prompt and input lyrics, in order to select the most accurate prediction from multiple Whisper trials, which is done in the first phase as illustrated in Section~\ref{sec:chatgpt}.
\cite{shum2023automatic} designs a gradient-guided strategy to select prompts. By contrast, we simply feed ChatGPT with an instruction to select prompts for itself.

c) 
The importance of a well-designed CoT~\cite{wei2022chain}, which effectively divides a complicated task into several phases and designs specific prompts for each phase, is widely acknowledged for enhancing LLM performance. We also proposes a concise CoT strategy, depicted in Section~\ref{sec:chatgpt}.

\section{Methodology}\label{sec:method}
The overall framework of our method is presented in~\figref{fig:framework}. 
This section will provide an in-depth analysis of the design of the Whisper and ChatGPT components, and our multilingual dataset.
 
\subsection{Whisper as Zero-shot Lyrics Transcriptor}\label{sec:whisper}
In the Whisper~\cite{radford2022robust} paper, the authors scaled the weakly supervised ASR to 680,000 hours of labeled audio data, which covers 96 languages and includes both multilingual and multitask training. This approach demonstrates high-quality results without the need for fine-tuning, self-supervision, or self-training techniques. By leveraging weak supervision and large-scale training, Whisper generalizes well to standard benchmarks and achieves robust speech recognition in various downstream tasks.

Motivated by this, we discovered that the weakly supervised Whisper model, trained on speech data, also excels in lyrics transcription within the music domain. We directly apply Whisper to transcribe lyrics of music from various genres, including pop, folk, rock, and rap, and find that the model consistently achieves accurate transcription results. The model excels at long-form transcription and is robust to different song styles, even for challenging genres such as rock and electronic music, where Whisper still provides reasonable results. We further test Whisper on multiple benchmark datasets for lyric transcription. The results indicate that Whisper, without any training or fine-tuning, can achieve or surpass SOTA performance across multiple lyric transcription datasets.

Upon analyzing the transcription results from Whisper, we observed that the model occasionally outputs content unrelated to lyrics, such as music descriptions, emojis, website watermarks, and YouTube advertisements. We attribute this to the weakly supervised training of Whisper on large-scale noisy speech datasets. To address this issue, we utilize the input prompt designed in Whisper as a prefix prompt to guide it toward the lyric transcription task. Unlike prompt designing philosophy in other large language models, Whisper's prefix prompt does not work well with explicit task instructions and has difficulty understanding lengthy explanations. In practice, we notice that using the simplest prompt, ``lyrics:", effectively prevents the model from outputting descriptions of the music in most cases, resulting in a significant improvement in transcription results. Therefore, in the following sections, this prompt is consistently used for Whisper's transcription input.

Additionally, we apply post-processing tricks to Whisper's output, utilizing the model's predicted no-speech probability to handle situations where predictions are made despite the absence of vocals in the song. Specifically, we drop predicted lines of lyrics with a no speech probability greater than $0.9$. This effectively filters out watermarks and advertisements, further enhancing the transcription results.

\begin{table}[tb!]
\centering
\scalebox{0.95}{
\begin{tabular}{|p{0.95\linewidth}|}
\hline
\small{\textbf{GPT-4 Instruction Prompt}}\\
\hline
\small{
Task:
As a GPT-4 based lyrics transcription post-processor, your task is to analyze multiple ASR model-generated versions of a song's lyrics and determine the most accurate version closest to the true lyrics. {\color{blue}Also filter out invalid lyrics when all predictions are nonsense.}

Input: 
The input is in JSON format: 

\text{\{``prediction\_1'': ``line1;line2;...'', ...\}}

Output:
Your output must be strictly in readable JSON format without any extra text:

\{

``reasons'': ``reason1;reason2;...'',

``closest\_prediction'': \text{<key\_of\_prediction>}

``output'': ``line1;line2...''

\}

Requirements:
For the "reasons" field, you have to provide a reason for the choice of the "closest\_prediction" field.
For the "closest\_prediction" field, choose the prediction key that is closest to the true lyrics. {\color{blue}Only when all predictions greatly differ from each other or are completely nonsense or meaningless, which means that none of the predictions is valid, fill in "None" in this field.}
For the "output" field, you need to output the final lyrics of closest\_prediction. {\color{blue}If the "closest\_prediction" field is "None", you should also output "None" in this field. The language of the input lyrics is English.}}\\
\hline
\end{tabular}
}
\caption{Instruction prompt for GPT-4 contextualized post-processing. We decompose this task into three consecutive phases, inspired by Chain-of-Thought prompting. Note that lines in blue indicate additional prompts used exclusively for multilingual dataset construction.}
\vspace{-4mm}
\label{tab:prompt}
\end{table}
\subsection{ChatGPT as Effective Lyrics Post-processor}\label{sec:chatgpt}

Although we addressed some issues with Whisper's predictions through prompt design and post-processing, we still cannot avoid transcription translation errors, as well as grammatical and syntactical errors. Furthermore, due to the inherently stochastic nature of temperature scheduling in Whisper, the transcription predictions vary with each run, leading to fluctuations in evaluation metrics. To reduce this variance and enhance overall accuracy, we generate $3$ to $5$ predictions for each input music under identical settings and employ ChatGPT as an expert in lyrics to ensemble these multiple predictions.

The crux of the problem lies in designing an effective prompt for ChatGPT to accomplish the ensemble task reasonably. As shown in Table~\ref{tab:prompt}, we first assign ChatGPT the role of a transcription post-processor, indicating that its task is to analyze multiple lyric transcription results and select the one it deems most accurate. We then stipulate that both input and output should be in JSON format to facilitate structured processing and provide detailed descriptions for each output field.

Drawing on the Chain-of-Thought in large language models for reasoning, we devised a concise thought chain for ChatGPT that decomposes lyrics post-processing into three consecutive phases. This involves first having ChatGPT analyze multiple lyric inputs and provide reasons for selection, then making a choice, and finally outputting the chosen lyric prediction. We test this approach using GPT-3.5 and the newly released GPT-4. The results demonstrate that using the analysis-selection-prediction prompt for ChatGPT's inference effectively enhances the final transcription results, with GPT-4 exhibiting a noticeably superior performance compared to GPT-3.5.

\subsection{Multilingual Lyrics Transcription Dataset}\label{sec:dataset}
\begin{table}
\small
    \centering
\scalebox{0.9}{
    \begin{tabular}{lcccc}
    \midrule
    \multirow{1}{*}{Dataset} & \multirow{1}{*}{Languages} & \multirow{1}{*}{Songs} &  \multirow{1}{*}{Lines} & \multirow{1}{*}{Duraion}\\ 
    \midrule
    DSing~\cite{dabike2019automatic} & 1 (en) & 4,324 & 81,092  & 149.1h \\ 
    MUSDB18~\cite{schulze2021phoneme}  & 1 (en) & 82    & 2,289    & 4.6h \\
    DALI-train~\cite{meseguer2020creating}   & 1 (en) & 3,913 & 180,034 & 208.6h  \\
    DALI-full~\cite{meseguer2020creating}   & 30\textbf{$^*$} & 5,358\textbf{$^*$} & - & -  \\
    \midrule
    MulJam (Ours) & 6 & 6,031 & 182,429 & 381.9h \\ 
    \bottomrule
    \end{tabular}
}
\caption{Comparison between different lyrics transcription datasets. 
Our model operates with a longer window (\textasciitilde$30$s), resulting in fewer lines compared to other datasets.
}
\vspace{-4mm}
\label{tab:dataset}
\end{table}
Building upon the exceptional performance of the proposed framework in lyric transcription tasks, we further extend it to the challenging task of multilingual lyric transcription, introducing the first large-scale, weakly supervised, and copyright-free multilingual lyric transcription dataset. We utilize the publicly available MTG-Jamendo dataset for music classification, which comprises 55,000 full audio tracks, 195 tags, and music in various languages.

Since the MTG dataset contains a considerable proportion of non-vocal music, we first employ PANNs~\cite{kong2020panns}, a large-scale pre-trained audio pattern recognition model, to detect audio events and filter out non-vocal music with vocal-related tag probabilities below a predefined threshold. This filtering method eliminates approximately $60\%$ of the music, thereby substantially reducing the time and resources required for dataset construction. We then utilize Whisper to transcribe lyrics from the music.

As the music in the MTG dataset encompasses multiple languages, we first utilize the Language Identification module within Whisper to predict the language of input music. Based on the predicted language, we translate the prefix prompt ``lyrics:'' into the corresponding language for input,~\emph{e.g.}, ``paroles'' in French, and ``liedtext'' in German. After obtaining the transcription results, we discard lyrics that are too short or too long. When ensembling the prediction results with ChatGPT, we also incorporate the language of lyrics as an input condition in the prompt. Given the prevalence of nonsensical content in the transcription results, we additionally require ChatGPT to evaluate the validity of the transcribed lyrics in the prompt. If all input lyrics are deemed nonsensical,~\emph{e.g.}, all special Unicode characters, or extremely divergent, the transcription result for that piece of music is considered invalid and discarded.

To prepare the dataset for training, it is essential to conduct line-level annotation. Timestamps can be obtained from the output of Whisper by aligning the lyrics both before and after ChatGPT processing. For the alignment of strings, the Levenshtein distance \cite{levenshtein1966binary} is employed. To exclude aligned lines of lower confidence, the distance is normalized, setting a threshold at 0.2. The quality of annotation is further enhanced through two subsequent filtering stages. In the first stage, lines that exhibit unusually high character rates, exceeding 37.5 Hz, are eliminated. The second stage encompasses another Whisper iteration; segments yielding a transcription of "Thank you." are excluded. These segments, which typically represent instrumental sections, are believed to originate from Whisper's training on data similar to video transcripts.

Following the construction process outlined above, we ultimately obtained a multilingual lyric transcription dataset, MulJam, consisting of 6,031 songs with 182,429 lines and a total duration of 381.9 hours. The dataset's statistical information and comparisons with existing ALT datasets are presented in Table~\ref{tab:dataset}. 

To our best knowledge, MulJam is the first publicly available large-scale dataset for multilingual lyrics transcription without copyright restrictions. While DALI~\cite{meseguer2019dali} is another large-scale music dataset featuring multilingual lyrics, its restricted access and strict licensing requirements limit its applicability for downstream tasks. In contrast, MulJam is free from copyright-related constraints and can be utilized without approval, as the audio can be legally downloaded directly from public sources without the need for approval, making it easily accessible.
This even includes audio that is permitted for use in the development of commercial software. 
Researchers are permitted to legally modify our dataset for derivative works and redistribution, provided they cite our work and adhere to the CC BY-NC-SA license. Furthermore, in contrast to the imbalanced language distribution in DALI, where English songs account for over $80\%$ of the total songs, our dataset includes a greater proportion of songs in other languages, which is advantageous for multilingual lyrics transcription.

\begin{table}
    \centering
    \begin{tabular}{lccc}
        \hline
        Method & Jamendo & Hansen & DSing \\
        \hline
        TDNN-F~\cite{dabike2019automatic}             & 76.37  & 77.59 & 19.60 \\
        CTDNN-SA~\cite{demirel2020automatic}           & 66.96  & 78.53 & 14.96 \\
        Genre-informed AM~\cite{gupta2020automatic}  & 50.64  & 39.00 & 56.90 \\
        MSTRE-Net~\cite{demirel2021mstre}          & 34.94  & 36.78 & 15.38 \\
        DE2-segmented~\cite{demirel2021low}      & 44.52  & 49.92 & -     \\
        W2V2-ALT~\cite{ou2022transfer}           & 33.13  & 18.71 & \textbf{12.99} \\
        \cellcolor{mygray-bg}LyricWhiz (Ours)  & \cellcolor{mygray-bg}\textbf{24.25}  & \cellcolor{mygray-bg}\textbf{7.85} & \cellcolor{mygray-bg} 13.78 \\
        \cellcolor{mygray-bg}\quad w/o ChatGPT Ens. & \cellcolor{mygray-bg}\underline{28.18}  & 
        \cellcolor{mygray-bg}\underline{8.07}  & 
        \cellcolor{mygray-bg}15.22 \\
        \cellcolor{mygray-bg}\quad \quad w/o Whis. Prompt  & \cellcolor{mygray-bg}33.21  & 
        \cellcolor{mygray-bg}8.75  & 
        \cellcolor{mygray-bg}\underline{13.40} \\
        \hline
    \end{tabular}
    \caption{The WERs (\%) of various ALT systems, including ablation methods, on multiple datasets. Note that W2V2-ALT is an in-domain baseline that natively train on DSing. The results of our method on Jamendo, Hansen are obtained from full-length transcription results, and the results on DSing are obtained from utterance-level segments.}
    \label{tab:exp}
    \vspace{-4mm}
\end{table}
\section{Experiments}\label{sec:exp}
In this section, we first outline our experimental setup, including datasets and evaluation metrics. Next, we report lyrics transcription results on various benchmark datasets. We also conduct extensive ablation studies to verify the effectiveness of our methods. Finally, we demonstrate the reliability of our dataset through noise level estimation.

\subsection{Experimental Setup}\label{sec:exp_cond}
\noindent\textbf{Datasets.}~
Our proposed method does not require any training; thus, we directly test it on several accessible lyric transcription benchmark datasets, including Jamendo~\cite{stoller2019end}, Hansen~\cite{hansen2012recognition}, MUSDB18~\cite{schulze2021phoneme}, DSing~\cite{dabike2019automatic}. Among these, Jamendo, Hansen, and DSing are widely used test datasets in music transcription. MUSDB18, originally a dataset for music source separation, contains 150 rock-pop songs. The authors in~\cite{schulze2021phoneme} provided line-level lyric annotations for MUSDB18, making it a challenging real-world dataset for lyric transcription. Additionally, we manually collected 40 multilingual songs with lyrics annotations from MTG-Jamendo as a test set for the proposed dataset, which can be used to validate the reliability of our proposed dataset via transcription accuracy.

\noindent\textbf{Evaluation.}~
We report the Word Error Rate (WER) as the evaluation metric, which is the ratio of the total number of insertions, substitutions, and deletions with respect to the total number of words. We calculate the average WER on the test sets. Since Whisper possesses the capability for long-form transcription, we directly evaluate entire songs using Jamendo, Hansen, and the multilingual test set. We perform utterance-level evaluations on MUSDB18 and DSing since they only have utterance-level annotations. We discovered that many songs in these evaluation datasets are problematic, such as incorrect lyric annotations and excessively short song segments. One notable problem is that sometimes there are prominent harmony parts in the background of a song. However, it is not provided in the lyric annotations (e.g., Adele's ``Rolling in the Deep''). LyricWhiz is powerful enough to transcript both the leading vocal and the background vocal with high accuracy. Therefore, we removed these problematic data from our evaluations. Finally, we normalize the transcription results to match the standardized ground truths. We remove all special Unicode characters, such as emojis. All text is converted to lowercase, and numeric characters are converted to their alphabetic correspondence.

\noindent\textbf{Budget.}~
To ensure fast and multi-round inference of the Whisper-large model on various datasets, including the large-scale MTG-Jamendo dataset, we conducted our experiments concurrently on a server with 8xA100 80G GPUs. It takes approximately 9 hours to complete one round of inference, and each process uses up to 12G VRAM. The vocal probability threshold is set to 0.07 for PANNs-based vocal event detection. To carry out contextualized post-processing using ChatGPT, we invested a total of US\$2,000 on GPT-4 API for the entire project.

\subsection{Comparative Experiments}
\begin{table}
    \centering
    \renewcommand{\arraystretch}{1.}
    \begin{tabular}{lccc}
        \hline
        Method & a) & b) & c) \\
        \hline
        CTDNN-SA-mixture~\cite{schulze2021phoneme}          & 76.06  & 78.44 & 89.24 \\
        \cellcolor{mygray-bg}Ours-mixture  & \cellcolor{mygray-bg}\textbf{50.90}  & \cellcolor{mygray-bg}\textbf{47.04} & \cellcolor{mygray-bg}\textbf{50.70} \\
        \midrule
        CTDNN-SA-vocals~\cite{schulze2021phoneme}           & 37.83  & 30.85 & 58.45 \\
        \cellcolor{mygray-bg}Ours-vocals  & \cellcolor{mygray-bg}\textbf{26.29}  & \cellcolor{mygray-bg}\textbf{25.27} & \cellcolor{mygray-bg}\textbf{33.30} \\
        \hline
    \end{tabular}
    \caption{The WERs (\%) of our method and baseline~\cite{schulze2021phoneme} on three subsets of annotated MUSDB18. The results of our method  are obtained from utterance-level segments.}
    \label{tab:musdb}
    \vspace{-4mm}
\end{table}

In order to verify the superiority of our approach, we compare it with several previous studies on benchmark datasets. W2V2-ALT~\cite{ou2022transfer}, a transfer learning method based on ASR self-supervised models, represents the current state-of-the-art in lyric transcription tasks. In our experiments, we primarily compare our method with W2V2-ALT, as well as other previous methods. The experimental results, as shown in Table~\ref{tab:exp}, indicate that our method achieves the best performance on Jamendo and Hansen and the second-best performance on DSing. In long-form transcription datasets such as Jamendo and Hansen, our method significantly outperforms all previous approaches due to the strong contextual memory capabilities of both Whisper and ChatGPT. Furthermore, our method also leads by a considerable margin on MUSDB18, shown in Table~\ref{tab:musdb}, demonstrating the robust performance and resilience of our proposed method in more diverse and complex musical scenarios. It is worth noting that our method did not surpass previous results on the DSing dataset, which we attribute to two factors. First, previous models were trained on the DSing training set, making the DSing test set an in-distribution dataset for the models, while our approach does not require any training and directly employs large-scale ASR models for zero-shot lyric transcription. Second, the segmented evaluation on DSing results in the loss of contextual information, which consequently leads to inaccurate transcriptions.

\subsection{Ablation Studies}
To further substantiate the efficacy of each component within our proposed approach, we conducted comprehensive ablation experiments.

\noindent\textbf{Whisper Prompt.}~
In our experiments, we investigate the Whisper prompt mechanism and test various prompts. First, we construct a complex prompt following the format of ChatGPT prompts, including task descriptions, format specifications, and specific requirements. We then gradually reduce the constituent elements of the prompt and observe the results. We discover that, unlike general large language models, Whisper has weaker task understanding capabilities for complex prompts and can only comprehend shorter task prompts. In practice, using the simplest prompt ``lyrics:'' yielded the best results. For multilingual transcription, we translate "lyrics:" into the corresponding language. As shown in Table~\ref{tab:exp}, the designed prompt performs better in long-form transcription scenarios, assisting the model in producing meaningful lyrics for difficult tasks. However, its performance is less effective at the utterance level, possibly because predicting a single line of lyrics does not require additional contextual information.

\noindent\textbf{ChatGPT Ensemble.}~
In order to confirm that ChatGPT can analyze and infer the most accurate version of lyrics, we first conduct a simple experiment. In this experiment, we add the ground truth lyrics to the predicted results and input them together into ChatGPT for ensembling. We then calculate the proportion of times ChatGPT ultimately chose the ground truth. If ChatGPT is able to choose the most accurate lyrics,~\emph{i.e.}, the ground truth, the final proportion should be close to $100\%$. The computed results on the Hansen dataset is $72.7\%$ for ground truth data, which is sufficient to demonstrate that ChatGPT can make correct choices based on the constructed prompt and input lyrics. As further observed in Table~\ref{tab:exp}, ChatGPT ensembling is particularly effective for long-form lyric transcription, suggesting that ChatGPT requires contextual information (the content of preceding and following lyrics, as well as the content of different versions of predicted lyrics) for inference. In contrast, utterance-level lyric inputs lack both context and diversity among different prediction results, leading to inferior performance.

\begin{table}
    \centering
    \renewcommand{\arraystretch}{1.1}
    \begin{tabular}{lccc}
        \hline
        Language & Songs$_{train}$ & Songs$_{test}$ & WER$_{test}$  \\
        \hline
        English  & 3,791  & 20 & 21.86 \\
        French   & 1,030  & 7  & 26.64 \\
        Spanish  & 620    & 5  & 22.54 \\
        Italian  & 311    & 3  & 44.01 \\
        Russian  & 147    & 4  & 39.18 \\
        German   & 132    & 1  & 25.43 \\
        \midrule
        Overall  & 6,031  & 40 & 26.26\\
        \hline
    \end{tabular}
    \caption{The distribution of our dataset and WERs (\%) on test set. We manually constructed a test set of 40 songs following the language distribution of the collected training set. Then, we applied our proposed method to the test set and computed the WER.}
    \label{tab:noise}
    \vspace{-4mm}
\end{table}
\subsection{Dataset Analysis}
In order to demonstrate the reliability of the dataset constructed using Whisper and ChatGPT on MTG-Jamendo, we manually create a multilingual test set for noise level estimation. Specifically, we first select six languages from the intersection of the languages in MTG and those in which Whisper performs best. We then conduct a stratified sampling of 40 songs on Jamendo and manually annotate their lyrics. We use these 40 songs as a test set, assessing the WER to estimate the noise level of our collected dataset. Table~\ref{tab:noise} presents the number of songs in each language and the WER results for the test set, where our method achieves decent WER levels for the majority of languages. As our goal is to construct a large-scale, multilingual dataset for weak supervision, our method's transcription results are acceptable. Furthermore, we have not implemented specific normalization for multilingual transcription results, such as removing diacritical marks, which could be employed to enhance performance.

\section{Conclusion}\label{sec:conclusion}

This paper presents LyricWhiz, a novel zero-shot automatic lyrics transcription system excelling in various datasets and music genres. Combining Whisper and GPT-4, our approach significantly reduces WER in English and efficiently transcribes multiple languages. LyricWhiz further generates the first publicly accessible, large-scale, multilingual lyrics dataset with a human-annotated subset for noise level estimation and evaluation. 
The successful integration of the large speech model and large language model in LyricWhiz offers a novel avenue for traditional Music Information Retrieval (MIR) tasks, as previous task-specific solutions are being eclipsed by general-purpose models. Notably, large language models have demonstrated their superior language understanding abilities across various tasks. Hence, we anticipate further applications of large language models to a broader spectrum of music-related domains, such as text-to-music generation, to enhance the performance of various models.

\section{Acknowledgements}

We gratefully acknowledge the dataset post-processing work described in Section~\ref{sec:dataset} offered by Jiawen Huang. Jiahao Pan and Wei Xue were supported by the Theme-based Research Scheme (T45-205/21-N) and Early Career Scheme (ECS-HKUST22201322), Research Grants Council of Hong Kong. Yinghao Ma is a research student at the UKRI Centre for Doctoral Training in Artificial Intelligence and Music, supported by UK Research and Innovation [grant number EP/S022694/1]. Yizhi Li is fully funded by an industrial PhD studentship (Grant number: 171362) from the University of Sheffield, UK.

\bibliography{ISMIR}

\end{document}